\documentclass[10pt,twocolumn,letterpaper]{article}

\usepackage{iccv}
\usepackage{times}
\usepackage{epsfig}
\usepackage{graphicx}
\usepackage{amsmath}
\usepackage{amssymb}

\usepackage{booktabs}
\usepackage{multirow}
\usepackage{footmisc}
\usepackage{array}
\usepackage{tabularx}
\usepackage[dvipsnames]{xcolor}
\DeclareMathOperator*{\Softmax}{Softmax}
\DeclareMathOperator*{\ReLU}{ReLU}

\usepackage[pagebackref=true,breaklinks=true,colorlinks,bookmarks=false]{hyperref}

\iccvfinalcopy 

\def\iccvPaperID{1119} 

\begin{document}

\title{Language-Conditioned Graph Networks for Relational Reasoning}

\author{Ronghang Hu$^1$~~~~~~Anna Rohrbach$^1$~~~~~~Trevor Darrell$^1$~~~~~~Kate Saenko$^2$ \\
$^1$University of California, Berkeley $\qquad$ $^2$Boston University \\
}

\maketitle

\begin{abstract}
Solving grounded language tasks often requires reasoning about relationships between objects in the context of a given task. For example, to answer the question ``What color is the mug on the plate?'' we must check the color of the specific mug that satisfies the ``on'' relationship with respect to the plate. Recent work has proposed various methods capable of complex relational reasoning. However, most of their power is in the inference structure, while the scene is represented with simple local appearance features. In this paper, we take an alternate approach and build contextualized representations for objects in a visual scene to support relational reasoning. We propose a general framework of Language-Conditioned Graph Networks (LCGN), where each node represents an object, and is described by a context-aware representation from related objects through iterative message passing conditioned on the textual input. E.g., conditioning on the ``on'' relationship to the plate, the object ``mug'' gathers messages from the object ``plate'' to update its representation to ``mug on the plate'', which can be easily consumed by a simple classifier for answer prediction. We experimentally show that our LCGN approach effectively supports relational reasoning and improves performance across several tasks and datasets. Our code is available at \url{http://ronghanghu.com/lcgn}.
\vspace{-1em}

\end{abstract}

\section{Introduction}
\label{sec:intro}

\begin{figure}[t]
\vspace{-1em}
\centering
\includegraphics[width=.95\columnwidth]{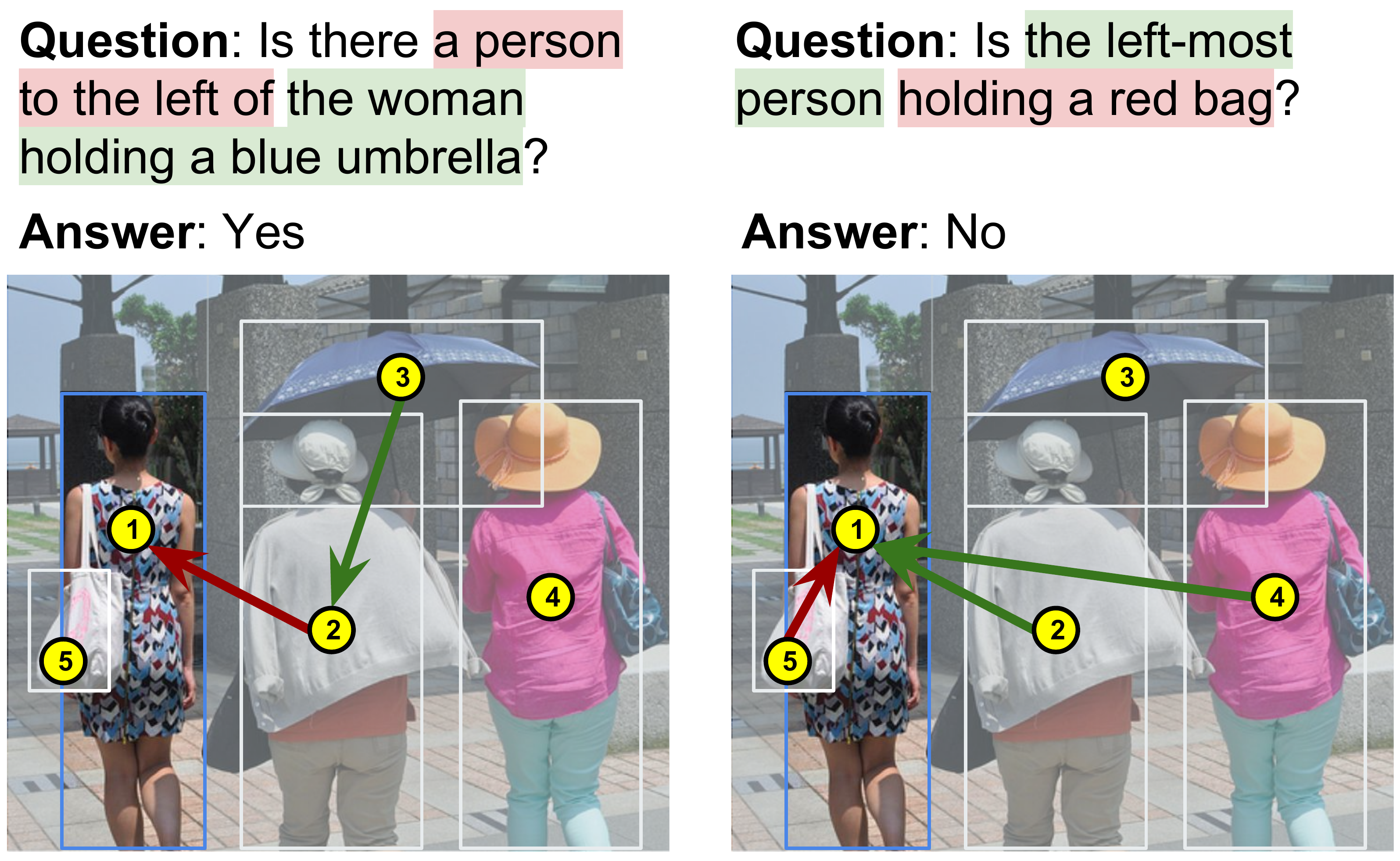}
\caption{\small
In this work, we create context-aware representations for objects by sending messages between relevant objects in a dynamic way that depends on the input language. In the left example, the first round of message passing updates object 2 with features of object 3 based on \textit{the woman holding a blue umbrella} (green arrow), and the second round updates object 1 with object 2's features based on \textit{person to the left} (red arrow). The final answer prediction can be made by a single attention hop over the most relevant object (blue box).}
\label{fig:teaser}
\vspace{-1.5em}
\end{figure}

Grounded language comprehension tasks, such as visual question answering (VQA) or referring expression comprehension (REF), require finding the relevant objects in the scene and reasoning about certain relationships between them. For example in Figure~\ref{fig:teaser}, to answer the question \textit{is there a person to the left of the woman holding a blue umbrella}, we must locate the relevant objects -- \textit{person}, \textit{woman} and \textit{blue umbrella} -- and model the specified relationships -- \textit{to the left of} and \textit{holding}.

How should we build a model to perform reasoning in  grounded language comprehension tasks? Prior works have explored various approaches from learning joint visual-textual representations (\eg \cite{fukui2016multimodal,perez2018film}) to pooling over pairwise relationships (\eg \cite{santoro2017simple,yu2018mattnet}) or constructing explicit reasoning steps with modular or symbolic representations (\eg \cite{andreas16neural,yi2018neural}). Although these models are capable of performing complex relational inference, their scene representations are built upon local visual appearance features that do not contain much contextual information. Instead, they tend to rely heavily on manually designed inference structures or modules to perform reasoning about relationships, and are often specific to a particular task. 
 
In this work, we propose an alternative way to facilitate reasoning with a context-aware scene representation, suitable for multiple tasks. Our proposed Language-Conditioned Graph Network (LCGN) model augments the local appearance feature of each entity in the scene with a relational contextualized feature. Our model is a graph network built upon visual entities in the scene, which collects relational information through multiple iterations of message passing between the entities. It dynamically determines which objects to collect information from on each round, by weighting the edges in the graph, and sends messages through the graph to propagate just the right amount of relational information. The key idea is to condition the message passing on the specific contextual relationships described in the input text. Figure~\ref{fig:teaser} illustrates this process, where the \textit{person} would be represented not only by her local appearance, but also by contextualized features indicating her relationship to other relevant objects in the scene, e.g., \textit{left of a woman}.  
Our contextualized representation can be easily plugged into task-specific models to replace standard local appearance features, facilitating reasoning with rich relational information. E.g. for the question answering task, it is sufficient to perform a single attention hop over the relevant object, whose representation is contextualized (e.g. blue box in Figure~\ref{fig:teaser}). 

Importantly, our scene representation is constructed with respect to the given reasoning task. An object in the scene may be involved in multiple relations in different contexts: in Figure~\ref{fig:teaser}, the person can be simultaneously \textit{left of a woman holding a blue umbrella}, \textit{holding a white bag}, and \textit{standing on a sidewalk}. Rather than building a complete representation of all the first- and higher-order relational information for each object (which can be enormous and unnecessary), we focus the contextual representation on relational information that is helpful to the reasoning task by conditioning on the input text (Figure~\ref{fig:teaser} left vs. right).

We apply our Language-Conditioned Graph Networks to two reasoning tasks with language inputs---Visual Question Answering (VQA) and Referring Expression Comprehension (REF). 
In these tasks, we replace the local appearance-based visual representations with the context-aware representations from our LCGN model, and demonstrate that our context-aware scene representations can be used as inputs to perform complex reasoning via simple task-specific approaches, with a consistent improvement over the local appearance features across different tasks and datasets. We obtain state-of-the-art results on the GQA dataset \cite{hudson2019gqa} for VQA and the CLEVR-Ref+ dataset \cite{liu2019clevr} for REF. 

\section{Related work}
\label{sec:related}

We first provide an overview of the reasoning tasks addressed in this paper. Then we review related work on graph networks and other contextualized representations. Finally, we discuss alternative approaches to reasoning problems.

\vspace{-1em}
\paragraph{Visual question answering (VQA) and referring expression comprehension (REF)} VQA and REF are two popular tasks that require reasoning about image content. While in VQA the goal is to answer a question about an image~\cite{antol2015vqa}, in REF one has to localize an image region that corresponds to a referring expression~\cite{mao2016generation}. While the real-world VQA dataset~\cite{antol2015vqa,goyal2017making} focuses more on perception than complex reasoning, the more recent synthetic CLEVR~\cite{johnson2017clevr} dataset is a standard benchmark for relational reasoning. An even more recent GQA dataset~\cite{hudson2019gqa} brings together the best of both worlds: real images and relational questions. It is built upon the Visual Genome dataset~\cite{krishna2017visual} and construct the balanced question-answer pairs from scene graphs.

For REF, there are a number of standard benchmarks such as RefCOCO \cite{yu2016modeling} and RefCOCOg \cite{mao2016generation}, with natural language referring expressions and images from the COCO dataset \cite{lin2014microsoft}. However, many of the expressions in these datasets do not require resolving relations. Recently, a new CLEVR-Ref+ dataset~\cite{liu2019clevr} has been proposed for REF. It is built using the CLEVR environment and involves very complex queries, aiming to assess the reasoning capabilities of existing models and find their limitations.

In this work we tackle both VQA and REF tasks on three datasets in total. Notably, in all cases, we use the same approach, Language-Conditioned Graph Network (LCGN), to build contextualized representations of objects/image regions. This shows the generality and effectiveness of our approach for various visual reasoning tasks.

\vspace{-1em}
\paragraph{Graph networks and contextualized representations}

Graph networks are powerful models that can perform relational inference through message passing~\cite{battaglia2018relational,gilmer2017neural,kipf2016semi,li2016gated,velivckovic2017graph,zhou2018graph}. The core idea is to enable communication between image regions to build contextualized representations of these regions. Graph networks have been successfully applied to various tasks, from object detection~\cite{liu2018structure} and region classification~\cite{chen2018iterative} to human-object interaction~\cite{qi2018learning} and activity recognition~\cite{herzig2018classifying}. Besides, self-attention models~\cite{vaswani2017attention} and non-local networks~\cite{wang2018non} can also be cast as graph networks in a general sense. 
Below we review some of the recent works that rely on graph networks and other contextualized representations for VQA and REF.

A prominent work that introduced relational reasoning in VQA is \cite{santoro2017simple}, which proposes Relation Networks (RNs) for modeling relations between all pairs of objects, conditioned on a question.  \cite{chang2018broadcasting} extends RNs with the Broadcasting Convolutional Network module, which globally broadcasts objects' visuo-spatial features. 
The first work to use graph networks in VQA is \cite{teney2017graph}, which combines dependency parses of questions and scene graph representations of abstract scenes. 
\cite{zhu2017structured} proposes modeling structured visual attention over a Conditional Random Field on image regions.
A recent work, \cite{norcliffe2018learning}, conditions on a question to learn a graph representation of an image, capturing object interactions with the relevant neighbours via spatial graph convolutions. 
Later, \cite{cadene2019murel} extends this idea to modeling spatial-semantic pairwise relations between all pairs of regions.

For the REF task, \cite{wang2019neighbourhood} proposes Language-guided Graph Attention Networks, where attention over nodes and edges is guided by a referring expression, which is decomposed into subject, intra-class and inter-class relationships.

Our work is related to, yet distinct from, the approaches above. 
While \cite{norcliffe2018learning} predicts a sparsely connected graph (conditioned on the question) that remains fixed for each step of graph convolution, our LCGN model predicts dynamic edge weights to focus on different connections in each message passing iteration. Besides,  \cite{norcliffe2018learning} is tailored to VQA and is non-trivial to adapt to REF (since it includes max-pooling over node representations). 
Compared to \cite{cadene2019murel}, instead of max-pooling over explicitly constructed pairwise vectors, our model predicts normalized edge weights that both improve computation efficiency in message passing and make it easier to visualize and inspect connections.
Finally, \cite{wang2019neighbourhood} is tailored to REF by modeling specific subject attention and inter- and intra-class relations, and does not gather higher-order relational information in an iterative manner. We propose a more general approach for scene representation that is applicable to both VQA and REF.

\vspace{-1em}
\paragraph{Reasoning models} 

A multitude of approaches have been recently proposed to tackle visual reasoning tasks, such as VQA and REF. Neural Module Networks (NMNs)~\cite{andreas16neural,hu2017learning} are multi-step models that build question-specific layouts and execute them against an image. NMNs have also been applied to REF, \eg CMN \cite{hu2017modeling} and Stack-NMN \cite{hu2018explainable}. MAC~\cite{hudson2018compositional} performs multi-step reasoning while recording information in its memory. 
FiLM~\cite{perez2018film} is an approach which modulates image representation with the given question via conditional batch normalization, and is extended in \cite{yao2018cascaded} with a multi-step reasoning procedure where both modalities can modulate each other. QGHC \cite{gao2018question} predicts question-dependent convolution kernels to modulate visual features. DFAF \cite{gao2019dynamic} introduces self-attention and co-attention mechanisms between visual features and question words, allowing information to flow across modalities. The Neural-Symbolic approach~\cite{yi2018neural} disentangles reasoning from image and language understanding, by first extracting symbolic representations from images and text, and then executing symbolic programs over them. 
MAttNet~\cite{yu2018mattnet}, a state-of-the-art approach to REF, uses attention to parse an expression and ground it through several modules. 

Our approach is not meant to substitute the aforementioned reasoning models, but to complement them. Our contextualized visual representation can be combined with other reasoning models to replace the local feature representation. 
A prominent reasoning model capable of addressing both VQA and REF is Stack-NMN~\cite{hu2018explainable}, and we empirically compare to it in Section~\ref{sec:exp}. 

\section{Language-Conditioned Graph Networks}
\label{sec:method}

Given a visual scene and a textual input for a reasoning task such as VQA or REF, we propose to construct a contextualized representation for each entity in the scene that contains the relational information needed for the reasoning procedure specified in the language input. 

This contextualized representation is obtained in our novel Language-Conditioned Graph Networks (LCGN) model, through iterative message passing conditioned on the language input. It can be then used as input to a task-specific output module such as a single-hop VQA classifier.

\subsection{Context-aware scene representation}
\label{sec:method_model}

\begin{figure*}[t]
\vspace{-1.5em}
\centering
\includegraphics[width=.95\textwidth]{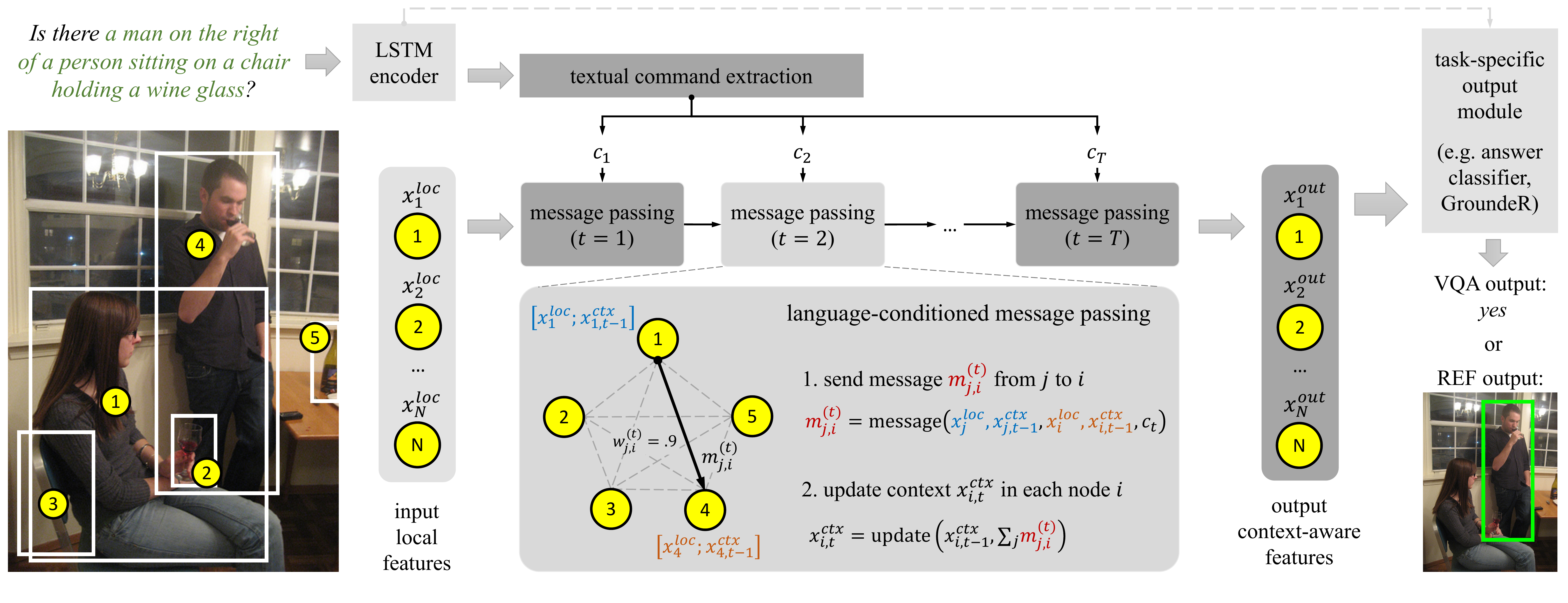}
\vspace{-0.5em}
\caption{We propose Language-Conditioned Graph Networks (LCGN) to address reasoning tasks such as VQA and REF. Our model constructs a context-aware representation $x^{out}_i$ for each object $i$ through iterative message passing conditioned on the input text. During message passing, each object $i$ is represented by a local feature $x^{loc}_i$ and a context feature $x^{ctx}_{i,t}$. In every iteration, each object $j$ sends a message vector $m^{(t)}_{j,i}$ to each object $i$, which is collected by $i$ to update its context feature $x^{ctx}_{i,t}$. The local feature $x^{loc}_i$ and the final context feature $x^{ctx}_{i,T}$ are combined into a joint context-aware feature $x^{out}_i$, which is used in simple task-specific output modules for VQA or REF.}
\label{fig:method}
\vspace{-1em}
\end{figure*}

For an image $I$ and a textual input $Q$ that represents a reasoning task, let $N$ be the number of entities in the scene, where each entity can be a detected object or a spatial location on the convolutional feature map of the image. Let $x^{loc}_i$ (where $i = 1, ..., N$) be the local feature representation of the $i$-th entity, \ie the $i$-th detected object's visual feature or the convolutional feature at the $i$-th location on the feature grid. We would like to output a context-aware representation $x^{out}_i$ 
for each entity $i$ conditioned on the textual input $Q$ that contains the relational context associated with entity $i$. This is obtained through iterative message passing over $T$ iterations with our Language-Conditioned Graph Networks, as shown in Figure~\ref{fig:method}.

We use a fully-connected graph over the scene, where each node corresponds to an entity $i$ as defined above, and there is a directed edge $i \rightarrow j$ between every pair of entities $i$ and $j$. 
Each node $i$ is represented by a local feature $x^{loc}_i$ that is fixed during message passing, and a context feature $x^{ctx}_{i,t}$ that is updated during each iteration $t$. A learned parameter is used as the initial context representation $x^{ctx}_{i,0}$ at $t=0$ for all nodes, before the message passing starts.

\vspace{-1em}
\paragraph{Textual command extraction} To incorporate the textual input in the iterative message passing, we build a textual command vector for each iteration $t$ (where $t = 1, ..., T$). Given a textual input $Q$ for the reasoning task, such as a question in VQA or a query in REF, we extract a set of vectors $\{c_t\}$ from the text $Q$, using the same multi-step textual attention mechanism as in Stack-NMN~\cite{hu2018explainable} and MAC~\cite{hudson2018compositional}. Specifically, $Q$ is encoded into a sequence $\{h_s\}_{s=1}^S$ and a summary vector $q$ with a bi-directional LSTM as: 
\begin{equation}
[h_1, h_2, ..., h_S] = \mathrm{BiLSTM}(Q) \quad \text{and} \quad q = \left[h_1; h_S\right]
\label{eqn:q_enc}
\end{equation}
where $S$ is the number of words in $Q$, and $h_s = [\overrightarrow{h_s}; \overleftarrow{h_s}]$ is the concatenation of the forward and backward hidden states for word $s$ from the bi-directional LSTM output. At each iteration $t$, a textual attention $\alpha_{t,s}$ is computed over the words, and the textual command $c_t$ is obtained from the textual attention as follows:
\begin{eqnarray}
    \alpha_{t,s} = \Softmax_s \left( W_1 \left( h_s \odot \left(W_2^{(t)} \ReLU\left(W_3 q\right) \right) \right) \right) \\
    c_t = \sum_{s=1}^S \alpha_{t,s} \cdot h_s \qquad\qquad\qquad\qquad\qquad\qquad\qquad \label{eqn:c_t}
\end{eqnarray}
where $\odot$ is element-wise multiplication. Each $c_t$ can be seen as a textual command supplied during the $t$-th iteration. Unlike all other parameters that are shared across iterations, here $W_2^{(t)}$ is learned separately for each iteration $t$. 

\vspace{-1em}
\paragraph{Language-conditioned message passing} At the $t$-th iteration where $t=1,...,T$, we first build a joint representation of each entity. Then, we compute the (directed) connection weights $w^{(t)}_{j,i}$ from every entity $j$ (the sender, $j=1,...,N$) to every entity $i$ (the receiver, $i=1,...,N$). Finally, each entity $j$ sends a message vector $m^{(t)}_{j,i}$ to each entity $i$, and each entity $i$ sums up all of its incoming messages $m^{(t)}_{j,i}$ to update its contextual representation from $x^{ctx}_{i,t-1}$ to $x^{ctx}_{i,t}$ as described below.

Step 1. We build a joint representation $\tilde{x}_{i,t}$ for each node, by concatenating $x^{loc}_i$ and $x^{ctx}_{i,t-1}$ and their element-wise product (after linear mapping) as
\begin{equation}
    \tilde{x}_{i,t} = \left[x^{loc}_i; x^{ctx}_{i,t-1}; \left( W_4 x^{loc}_i \right) \odot \left( W_5 x^{ctx}_{i,t-1} \right)\right]
\end{equation}

Step 2. We compute the directed connection weights $w_{j,i}^{(t)}$ from node $j$ (the sender) to node $i$ (the receiver), conditioning on the textual command $c_t$ at iteration $t$. Here, the connection weights are normalized with a softmax function over $j$, so that the sender weights sum up to $1$ for each receiver, \ie $\sum_{j=1}^N w_{j,i}^{(t)} = 1$ for all $i = 1,..., N$ as follows:
\begin{equation}
    w_{j,i}^{(t)} = \Softmax_j \left( \left(W_6 \tilde{x}_{i,t}\right)^T \left( \left(W_7 \tilde{x}_{j,t}\right) \odot \left(W_8 c_t\right) \right) \right) \label{eqn:edge_w}
\end{equation}

Step 3. Each node $j$ sends a message $m_{j,i}^{(t)}$ to each node $i$ conditioning on the textual input $c_t$ and weighted by the connection weight $w_{j,i}^{(t)}$. Then, each node $i$ sums up the incoming messages and updates its context representation:
\begin{eqnarray}
    m_{j,i}^{(t)} &=& w_{j,i}^{(t)} \cdot \left( \left(W_9 \tilde{x}_{j,t}\right) \odot \left(W_{10} c_t\right) \right) \label{eqn:msg} \\
    x^{ctx}_{i,t} &=& W_{11} \left[x^{ctx}_{i,t-1}; \sum_{j=1}^N m_{j,i}^{(t)}\right] \label{eqn:msg_sum}
\end{eqnarray}

A naive implementation would involve $N^2$ pairwise vectors $m_{j,i}^{(t)}$, which is inefficient for large $N$. We implement it more efficiently by building an $N$-row matrix $M$ containing $N$ unweighted messages $\tilde{m}^{(t)}_{j} = (W_9 \tilde{x}_{j,t}) \odot (W_{10} c_t)$ in Eqn.~\ref{eqn:msg}, which is left multiplied by the edge weight matrix $E$ (where $E_{ij}=w_{j,i}^{(t)}$) to obtain the sums $\sum_{j=1}^N m_{j,i}^{(t)}$ in Eqn.~\ref{eqn:msg_sum} for all nodes in a single matrix multiplication. With this implementation, we can train our LCGN model efficiently with $N$ as large as 196 in our experiments.

\vspace{-1em}
\paragraph{Final representation} We combine each entity's local feature $x^{loc}_i$ and context feature $x^{ctx}_{i,T}$ (after $T$ iterations) as its final representation $x^{out}_i$:
\begin{equation}
    x^{out}_i = W_{12} \left[x^{loc}_i; x^{ctx}_{i,T}\right]
\end{equation}

The $x^{out}_i$ can be used as input to subsequent task-specific modules such as VQA or REF models, instead of the original local representation $x^{loc}_i$.

\subsection{Application to VQA and REF}
\label{sec:method_app}

To apply our LCGN model to language-based reasoning tasks such as Visual Question Answering (VQA) and Referring Expression Comprehension (REF), we build simple task-specific output modules based on the language input and the contextualized representation of each entity. Our LCGN model and the subsequent task-specific modules are jointly trained end-to-end.

\vspace{-1em}
\paragraph{A single-hop answer classifier for VQA} The VQA task requires outputting an answer for an input image $I$ and a question $Q$. We adopt the commonly used classification approach and build a single-hop attention model as a classifier to select one of the possible answers from the training set.

First, the question $Q$ is encoded into a vector $q$ with the Bi-LSTM in Eqn.~\ref{eqn:q_enc}. Then a single-hop attention $\beta_i$ is used over the objects to aggregate visual information, which is fused with $q$ to predict the score vector $y$ for each answer.
\begin{eqnarray}
    \beta_i &=& \Softmax_i \left( W_{13} \left(x^{out}_i \odot \left(W_{14}q\right)\right) \right) \label{eqn:vqa_in} \\
    y &=& W_{15} \ReLU \left(W_{16} \left[\sum_{i=1}^N \beta_i x^{out}_i ; q \right]\right)
\end{eqnarray}
During training, a softmax or sigmoid classification loss is applied on the output scores $y$ for answer classification.

\vspace{-1em}
\paragraph{GroundeR \cite{rohrbach2016grounding} for REF} The REF task requires outputting a target bounding box as the grounding result for an input referring expression $Q$. Here, we use a retrieval approach as in previous works and select one target entity from the $N$ candidate entities in the scene (either object detection results or spatial locations on a convolutional feature map). To select the target object $p$ from the $N$ candidates, we encode expression $Q$ to vector $q$ as in Eqn~\ref{eqn:q_enc} and build a model similar to the fully-supervised version of GroundeR \cite{rohrbach2016grounding} to output a matching score $r_i$ for each entity $i$. 
In the case of using spatial locations on a convolutional feature map, we further output a $4$-dimensional vector $u$ to predict the bounding box offset from the feature grid location.
\begin{eqnarray}
    r_i &=& W_{17} \left(x^{out}_i \odot \left(W_{18}q\right)\right) \label{eqn:ref_in} \\
    p &=& \arg\max_i r_i \label{eqn:target_select} \\
    u &=& W_{19} x^{out}_p \label{eqn:bbox_reg}
\end{eqnarray}
During training, we use a softmax loss over the scores $r_i$ among the $N$ candidates to select the target entity $p$, and an L2 loss over the box offset $u$ to refine the box location.

\section{Experiments}
\label{sec:exp}

We apply our LCGN model to two tasks -- VQA and REF -- for language-conditioned reasoning. For the VQA task, we evaluate on the GQA dataset \cite{hudson2019gqa} and the CLEVR dataset \cite{johnson2017clevr}, which both require resolving relations between objects. For the REF task, we evaluate on the CLEVR-Ref+ dataset \cite{liu2019clevr}. In particular, the CLEVR and CLEVR-Ref+ datasets contain many complicated questions or expressions with higher-order relations, such as \textit{the ball on the left of the object behind a blue cylinder}.

\subsection{Visual Question Answering (VQA)}
\label{sec:exp_vqa}

\paragraph{Evaluation on the GQA dataset}
We first evaluate our LCGN model on the GQA dataset \cite{hudson2019gqa} for visual question answering. The GQA dataset is a large-scale visual question answering dataset with real images from the Visual Genome dataset \cite{krishna2017visual} and balanced question-answer pairs. Each training and validation image is also associated with scene graph annotations describing the classes and attributes of those objects in the scene, and their pairwise relations. Along with the images and question-answer pairs, the GQA dataset provides two types of pre-extracted visual features for each image -- convolutional grid features of size $7 \times 7 \times 2048$ extracted from a ResNet-101 network \cite{he2016deep} trained on ImageNet, and object detection features of size $N_{det} \times 2048$ (where $N_{det}$ is the number of detected objects in each image with a maximum of 100 per image) from a Faster R-CNN detector \cite{ren2015faster}. 

We apply our LCGN model together with the single-hop classifier (``\textbf{single-hop + LCGN}'') in Sec.~\ref{sec:method_app} for answer prediction. We use $T = 4$ rounds of message passing in our LCGN model, which takes approximately 20 hours to train using a single Titan Xp GPU. As a comparison to our LCGN model, we also train the single-hop classifier with only the local features $x^{loc}$ in Eqn.~\ref{eqn:vqa_in} (``\textbf{single-hop}''). 

\begin{table}[t]
\small
\vspace{-1.5em}
\begin{center}
\begin{tabular}{lccc}
\toprule
\multirow{2}{*}{Method} & \multicolumn{3}{c}{Accuracy\footref{fn:gqa_val}} \\
& val & test-dev & test \\
\midrule
CNN+LSTM \cite{hudson2019gqa} & 49.2\% & -- & 46.6\% \\
Bottom-Up \cite{anderson2017bottom} & 52.2\% & -- & 49.7\% \\
MAC \cite{hudson2018compositional} & 57.5\% & -- & 54.1\% \\
\midrule
single-hop & 62.0\% & 53.8\% & 54.4\% \\
single-hop + LCGN (ours) & \textbf{63.9\%} & \textbf{55.8\%} & \textbf{56.1\%} \\
\bottomrule
\end{tabular}
\end{center}
\vspace{-0.7em}
\caption{VQA performance on the GQA dataset.\footref{fn:gqa_val}}
\label{tab:gqa}
\vspace{-1.5em}
\end{table}

We first experiment with using the released object detection features in the GQA dataset as our local features $x^{loc}$, which is shown in \cite{hudson2019gqa} to perform better than the convolutional grid features, and compare with previous works.\footnote{We learned from the GQA dataset authors that its \textit{test-dev} and \textit{test} splits were collected differently from its \textit{train} and \textit{val} splits, with a noticeable domain shift from \textit{val} to \textit{test-dev} and \textit{test}. We train on the \textit{train} split and report results on three GQA splits (\textit{val}, \textit{test-dev} and \textit{test}). The performance of previous work on  \textit{val} was obtained from the dataset authors.\label{fn:gqa_val}} Similar to MAC \cite{hudson2018compositional}, we initialize question word embedding from GloVe \cite{pennington2014glove} and maintain an exponential moving average of model parameters during training. To facilitate spatial reasoning, we concatenate the Faster R-CNN object detection features with their corresponding box coordinates.
The results are shown in Table~\ref{tab:gqa}. By comparing ``single-hop + LCGN'' with ``single-hop'' in the last two rows, it can be seen that our LCGN model brings around 2\% (absolute) improvement in accuracy, indicating that our LCGN model facilitates reasoning by replacing the local features $x^{loc}$ with the contextualized features $x^{out}$ containing rich relational information for the reasoning task. Figure~\ref{fig:gqa_vis} shows question answering examples from our model on this dataset.

We compare with three previous approaches in Table~\ref{tab:gqa}. CNN+LSTM \cite{hudson2019gqa} and Bottom-Up \cite{anderson2017bottom} are simple fusion approaches between the text and the image, using the released GQA convolutional grid features or object detection features respectively. The MAC model \cite{hudson2018compositional} is a multi-step attention and memory model with specially designed control, reading and writing cells, and is trained on the same object detection features as our model. Our approach outperforms the MAC model that performs multi-step inference, obtaining the state-of-the-art results on the GQA dataset.

\begin{table}[t]
\small
\vspace{-1.5em}
\begin{center}
\begin{tabular}{lccc}
\toprule
\multirow{2}{*}{Method} & \multirow{2}{*}{Local features} & \multicolumn{2}{c}{Accuracy} \\
& & val & test-dev \\
\midrule
single-hop & convolutional & 55.0\% & 48.6\% \\
single-hop + LCGN & grid features & \textbf{55.3\%} & \textbf{49.5\%} \\
\midrule
single-hop & object features & 62.0\% & 53.8\% \\
single-hop + LCGN & from detection & \textbf{63.9\%} & \textbf{55.8\%} \\
\midrule
single-hop & GT objects & 87.0\% & n/a \\
single-hop + LCGN & and attributes\footref{fn:gqa_sg} & \textbf{90.2\%} & n/a \\
\bottomrule
\end{tabular}
\end{center}
\vspace{-0.7em}
\caption{Ablation on different local features on the GQA dataset.}
\label{tab:gqa_feat}
\vspace{-1.5em}
\end{table}

We further apply our LCGN model to other types of local features, and experiment with using either the same $7\times 7 \times 2048$-dimensional convolutional grid features (where each $x_i^{loc}$ is a feature map location and $N=49$) as used in CNN+LSTM in Table~\ref{tab:gqa} or an ``oracle'' symbolic local representation at both training and test time, based on a set of ground-truth objects along with their class and attribute annotations (``GT objects and attributes'') in the scene graph data of the GQA dataset. In the latter setting with symbolic representation, we construct two one-hot vectors to represent each object's class and attributes, and concatenate them as each object's $x^{loc}_i$.\footnote{In this setting, we can only evaluate on the \textit{val} split with public scene graph annotations. We note that this is the only setting where we use the scene graphs in the GQA dataset. In all other settings, we only use the images and question-answer pairs to train our models. Also, our model does not rely on the GQA question semantic step annotations in any settings. \label{fn:gqa_sg}} The results are shown in Table~\ref{tab:gqa_feat}, where our LCGN model delivers consistent improvements over all three types of local feature representations.

\vspace{-1em}
\paragraph{Evaluation on the CLEVR dataset} We also evaluate our LCGN model on the CLEVR dataset \cite{johnson2017clevr}, a dataset for VQA with complicated relational questions, such as \textit{what number of other objects are there of the same size as the brown shiny object}. Following previous works, we use the $14 \times 14 \times 1024$ convolutional grid features extracted from the C4 block of an ImageNet-pretrained ResNet-101 network \cite{he2016deep} as the local features $x^{loc}$ on the CLEVR dataset (\ie each $x_i^{loc}$ is a feature map location and $N=196$).

Similar to our experiments on the GQA dataset, we apply our LCGN model together with the single-hop answer classifier and compare it with using only the local features in the answer classifier. We also compare with previous works that use only question-answer pairs as supervision (\ie without relying on the functional program annotations in \cite{johnson2017clevr}).

The results are shown in Table~\ref{tab:clevr}. It can be seen that the single-hop classifier only achieves 72.6\% accuracy when using the local convolutional grid features $x^{loc}$ (``\textbf{single-hop}''), which is unsurprising since the CLEVR dataset often involves resolving multiple and higher-order relations beyond the capacity of the single-hop classifier alone. However, when trained together with the context-aware representation $x^{out}$ from our LCGN model, this same single-hop classifier (``\textbf{single-hop + LCGN}'') achieves a significantly higher accuracy of 97.9\% comparable to several state-of-the-art approaches on this dataset, showing that our LCGN model is able to embed relational context information in its output scene representation $x^{out}$. Among previous works, Stack-NMN \cite{hu2018explainable} and MAC \cite{hudson2018compositional} rely on multi-step inference procedures to predict an answer. RN \cite{santoro2017simple} pools over all $N^2$ pairwise object-object vectors to collect relational information in a single step. FiLM \cite{perez2018film} modulates the batch normalization parameters of a convolutional network with the input question. NS-CL \cite{mao2018neuro} learns symbolic representations of the scene and uses quasi-logical reasoning. Except for Stack-NMN \cite{hu2018explainable}, most previous works are tailored to the VQA task, and it is non-trivial to apply them to other tasks such as REF, while our LCGN model provides a generic scene representation applicable to multiple tasks. Figure~\ref{fig:clevr_vis} shows question answering examples of our model.

\begin{table}[t]
\small
\vspace{-1.5em}
\begin{center}
\begin{tabular}{lc}
\toprule
Method & Accuracy \\
\midrule
Stack-NMN \cite{hu2018explainable} & 93.0\% \\
RN \cite{santoro2017simple} & 95.5\% \\
FiLM \cite{perez2018film} & 97.6\% \\
MAC \cite{hudson2018compositional} & 98.9\% \\
NS-CL \cite{mao2018neuro} & \textbf{99.2\%} \\
\midrule
single-hop & 72.6\% \\
single-hop + LCGN (ours) & 97.9\% \\
\bottomrule
\end{tabular}
\end{center}
\vspace{-0.7em}
\caption{VQA performance on the test split of the CLEVR dataset. We use $T=4$ rounds of message passing in our LCGN model.}
\label{tab:clevr}
\vspace{-1.5em}
\end{table}

We further experiment with varying the number $T$ of message passing iterations in our LCGN model. In addition, to isolate the effect of conditioning on textual inputs during message passing, we also train and evaluate a restricted version of LCGN without text conditioning (``\textbf{single-hop + LCGN w/o txt}''), by replacing the $c_t$'s from Eqn~\ref{eqn:c_t} with a vector of all ones. The results are shown in Table~\ref{tab:clevr_step}, where it can be seen that using multiple rounds of iterations ($T>1$) leads to a significant performance increase, and it is crucial to incorporate the textual information $c_t$ into the message passing procedure. This is likely because the CLEVR dataset involves complicated questions that need multi-step context propagation. In addition, it is more efficient to collect the specific relational context relevant to the input question, instead of building a scene representation with a complete and unconditional knowledge base of all relational information that any input questions can query from.

Given that multi-round message passing ($T>1$) works better than using only a single round ($T=1$), we further study whether it is beneficial to have dynamic connection weights $w^{(t)}_{j,i}$ in Eqn.~\ref{eqn:edge_w} that can be different in each iteration $t$ to allow an object $i$ to focus on different context objects $j$ in different rounds. As a comparison, we train a restricted version of LCGN with static connection weights $w_{j,i}$ (``\textbf{single-hop + LCGN w/ static $w_{j,i}$}''), where we only predict the weights $w^{(1)}_{j,i}$ in Eqn.~\ref{eqn:edge_w} for the first round $t=1$, and reuse it in all subsequent rounds (\ie setting $w^{(t)}_{j,i} = w^{(1)}_{j,i}$ for all $t>1$). From the last row of Table~\ref{tab:clevr_step} it can be seen that there is a performance drop when restricting to static connection weights $w_{j,i}$ predicted only in the first round, and we also observe a similar (but larger) drop for the REF task in Sec.~\ref{sec:exp_ref} and Table~\ref{tab:clevr_refplus}. This suggests that it is better to have dynamic connections during each iteration, instead of first predicting a fixed connection structure on which iterative message passing is performed (\eg \cite{norcliffe2018learning}).

\begin{table}[t]
\small
\vspace{-1.5em}
\begin{center}
\begin{tabular}{lcc}
\toprule
Method & Steps $T$ & Accuracy \\
\midrule
single-hop & n/a & 72.6\% \\
\midrule
single-hop + LCGN & $T=1$ & 94.0\% \\
single-hop + LCGN & $T=2$ & 94.5\% \\
single-hop + LCGN & $T=3$ & 96.4\% \\
single-hop + LCGN & $T=4$ & \textbf{97.9\%} \\
single-hop + LCGN & $T=5$ & 96.9\% \\
\midrule
single-hop + LCGN w/o txt & $T=4$ & 78.6\% \\
single-hop + LCGN w/ static $w_{j,i}$ & $T=4$ & 96.5\% \\
\bottomrule
\end{tabular}
\end{center}
\vspace{-0.7em}
\caption{Ablation on iteration steps $T$ and whether to condition on the text or have dynamic weights, on the CLEVR validation split.}
\label{tab:clevr_step}
\vspace{-1.5em}
\end{table}

\subsection{Referring Expression Comprehension (REF)}
\label{sec:exp_ref}

Our LCGN model provides a generic approach to building context-aware scene representations and is not restricted to a specific task such as VQA. We also apply our LGCN model to the referring expression comprehension (REF) task, where given a referring expression that describes an object in the scene, the model is asked to localize the target object with a bounding box. 

We experiment with the CLEVR-Ref+ dataset \cite{liu2019clevr}, which contains similar images as in the CLEVR dataset~\cite{johnson2017clevr} and complicated referring expressions requiring relation resolution. On the CLEVR-Ref+ dataset, we evaluate with the bounding box detection task in \cite{liu2019clevr}, where the output is a bounding box of the target object and there is only one single target object described by the expression. A localization is consider correct if it overlaps with the ground-truth box with at least 50\% IoU. Same as in our VQA experiments on the CLEVR dataset in Sec.~\ref{sec:exp_vqa}, here we also use the $14 \times 14 \times 1024$ convolutional grid features from ResNet-101 C4 block as our local features $x^{loc}$ (\ie each $x_i^{loc}$ is a feature map location and $N=196$), with $T=4$ rounds of message passing. The final target bounding box is predicted with a 4-dimensional bounding box offset vector $u$ in Eqn.~\ref{eqn:bbox_reg} from the selected grid location $p$ in Eqn.~\ref{eqn:target_select}.

We apply our LCGN model to build a context-aware representation $x^{out}$ conditioned on the input referring expression, which is used as input to our implementation of the GroundeR approach \cite{rohrbach2016grounding} (Sec.~\ref{sec:method_app}) for bounding box prediction (``\textbf{GroundeR + LCGN}''). As a comparison, we train and evaluate the GroundeR model without our context-aware representation (``\textbf{GroundeR}''), using local features $x^{loc}$ as inputs in Eqn.~\ref{eqn:ref_in}. Similar to our experiments on the CLEVR dataset for VQA in Sec.~\ref{sec:exp_vqa}, we also ablate our LCGN model with not conditioning on the input expression in message passing (``\textbf{GroundeR + LCGN w/o txt}'') or using static connection weights $w_{j,i}$ predicted from the first round (``\textbf{GroundeR + LCGN w/ static $w_{j,i}$}''). 

\begin{figure*}[t]
\centering
\small
\begin{tabularx}{\linewidth}{*{6}{c}}
~~~~~~~~input image & ~~~~~~~~~~~~~~~~~$t = 1$ & ~~~~~~~~~~~~~~~~~~~~~~~$t = 2$ & ~~~~~~~~~~~~~~~~~~~~~~$t = 3$ & ~~~~~~~~~~~~~~~~~~~~~~~$t = 4$ & ~~~~~~~~~single-hop attention $\beta_i$ \\
\hline
\end{tabularx}

\vspace{1mm}

question: \textit{is the fence in front of the elephant green and metallic?} ~ prediction: \textit{yes} ~ ground-truth: \textit{yes} \\
\includegraphics[width=\linewidth,trim={0 8mm 0 7mm},clip]{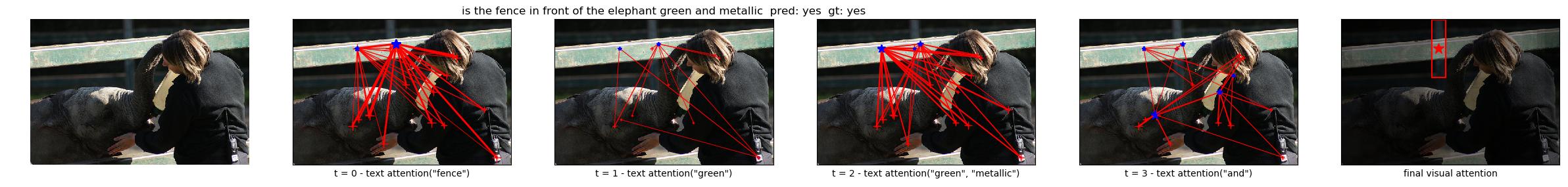} \\

question: \textit{the frisbee is on what animal?} ~ prediction: \textit{dog} ~ ground-truth: \textit{dog} \\
\includegraphics[width=\linewidth,trim={0 8mm 0 7mm},clip]{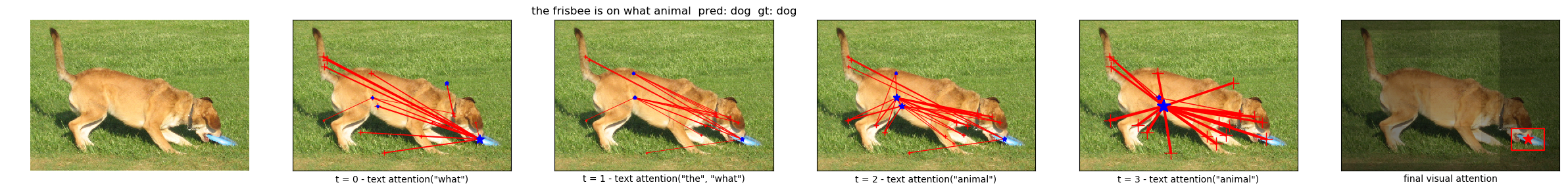} \\

\caption{Examples from our LCGN model on the validation split of the GQA dataset for VQA. In the middle 4 columns, each red line shows an edge $j \rightarrow i$ along the message passing paths (among the $N$ detected objects) where the connection edge weight $w^{(t)}_{j,i}$ exceeds a threshold. The \textcolor{blue}{blue} star on each line is the sender node $j$, and the line width corresponds to its connection weight. In the upper example, the person, the elephant and the fence propagate messages with each other, and fence receives messages from the elephant in $t=4$. In the lower example, the frisbee collect messages from the dog as contextual information in multiple rounds, and is picked up by the single-hop classifier. The \textcolor{red}{red} star (along with the box) in the last column shows the object with the highest single-hop attention $\beta_i$ in Eqn.~\ref{eqn:vqa_in}.}
\label{fig:gqa_vis}

\vspace{1em}

\centering
\small
\begin{tabularx}{\linewidth}{*{6}{c}}
~~~~~~~~input image & ~~~~~~~~~~~~~~~~~$t = 1$ & ~~~~~~~~~~~~~~~~~~~~~~~$t = 2$ & ~~~~~~~~~~~~~~~~~~~~~~$t = 3$ & ~~~~~~~~~~~~~~~~~~~~~~~$t = 4$ & ~~~~~~~~~single-hop attention $\beta_i$ \\
\hline
\end{tabularx}

\vspace{1mm}

question: \textit{what color is the matte ball that is the same size as the gray metal thing?} ~
prediction: \textit{yellow} ~ ground-truth: \textit{yellow} \\
\includegraphics[width=\linewidth,trim={0 8mm 0 7mm},clip]{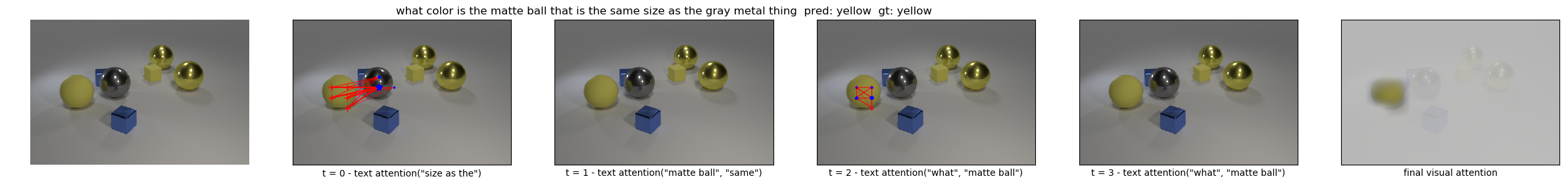} \\

question: \textit{how many other things are the same size as the yellow rubber ball?} ~ prediction: \textit{3} ~ ground-truth: \textit{3} \\
\includegraphics[width=\linewidth,trim={0 8mm 0 7mm},clip]{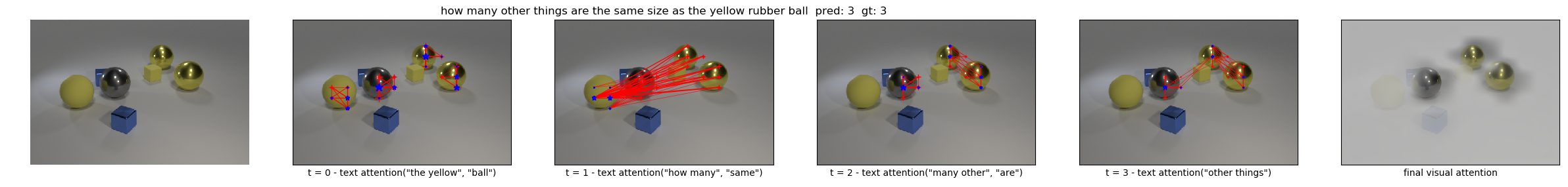} \\

\caption{Examples from our LCGN model on the validation split of the CLEVR dataset for VQA. The middle 4 columns show the connection edge weights $w^{(t)}_{j,i}$ similar to Figure~\ref{fig:gqa_vis}, where the \textcolor{blue}{blue} stars are the sender nodes. The last column shows the single-hop attention $\beta_i$ in Eqn.~\ref{eqn:vqa_in} over the $N = 14 \times 14$ feature grid. In the upper example, in $t=1$ the matte ball (leftmost) collects messages from the gray metal ball (of the same size), and then in $t=3$ messages are propagated within the convolutional grids on the matte ball, possibly to refine the collected context from the gray ball. In the lower example, in $t=1$ all four balls try to propagate messages within the convolutional grids of each ball region, and in $t=2$ the three other balls (of the same size) receive messages from the rubber ball (leftmost) and are picked up by the single-hop classifier.}
\label{fig:clevr_vis}

\vspace{1em}

\centering
\small
\begin{tabularx}{\linewidth}{*{6}{c}}
~~~~~~~~input image & ~~~~~~~~~~~~~~~~~$t = 1$ & ~~~~~~~~~~~~~~~~~~~~~~~$t = 2$ & ~~~~~~~~~~~~~~~~~~~~~~$t = 3$ & ~~~~~~~~~~~~~~~~~~~~~~~$t = 4$ & ~~~~~~~~~~bounding box output \\
\hline
\end{tabularx}

\vspace{1mm}

referring expression: \textit{any other things that are the same shape as the big matte thing(s)} \\
\includegraphics[width=\linewidth,trim={0 8mm 0 7mm},clip]{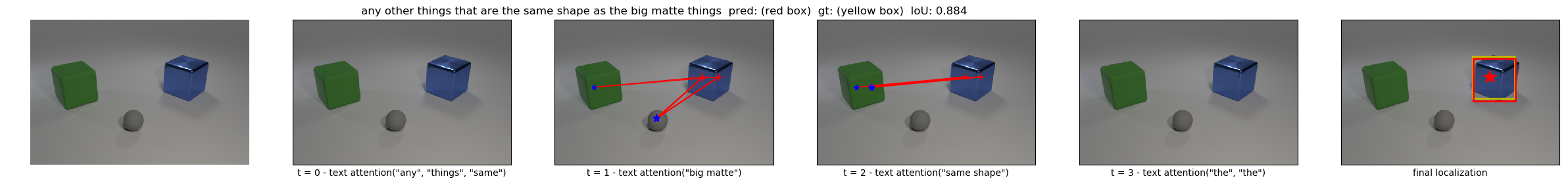} \\

referring expression: \textit{the second one of the cube(s) from right} \\
\includegraphics[width=\linewidth,trim={0 8mm 0 7mm},clip]{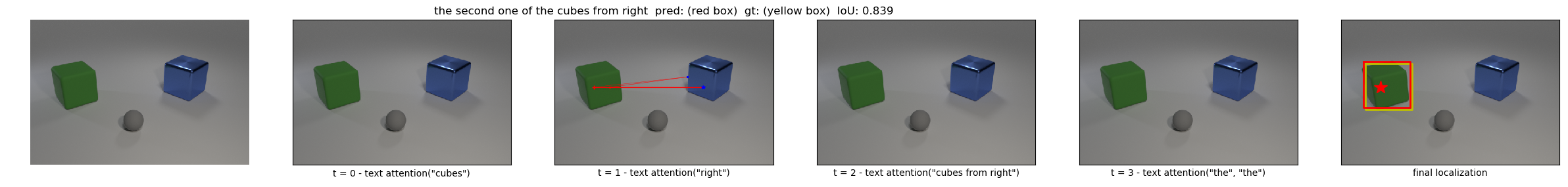} \\

\caption{Examples from our LCGN model on the validation split of the CLEVR-Ref+ dataset for REF. The middle 4 columns show the connection edge weights $w^{(t)}_{j,i}$ similar to Figure~\ref{fig:gqa_vis}, where the \textcolor{blue}{blue} stars are the sender nodes. The last column shows the selected target grid location $p$ on the $N = 14 \times 14$ spatial grid (the \textcolor{red}{red} star) in Eqn.~\ref{eqn:target_select}, along with the ground-truth (\textcolor[rgb]{.7,.7,0}{yellow}) box and the predicted box (\textcolor{red}{red} box from bounding box regression $u$ in Eqn.~\ref{eqn:bbox_reg}). In the upper example, the blue cube (the target object) collects messages from the two other objects in $t=2$, and then the blue cube further collects messages from the big matte green cube on the left (which has the same shape) in $t=3$. In the lower example, the green cube checks for other cubes by collecting messages from things on its right in $t=2$.}
\label{fig:clevr_refplus_vis}
\end{figure*}

The results are shown in Table~\ref{tab:clevr_refplus}, where our context-aware scene representation from LCGN leads to approximately 13\% (absolute) improvement in REF accuracy. Consistent with our observation on the VQA task, for the REF task we find it important for the message passing procedure to depend on the input expression, and allowing the model to have dynamic connection weights $w^{(t)}_{j,i}$ that can differ for each round $t$. Our model outperforms previous work by a large margin, achieving the state-of-the-art performance for REF on the CLEVR-Ref+ dataset. Figure~\ref{fig:clevr_refplus_vis} shows example predictions of our model on the CLEVR-Ref+ dataset. 

\begin{table}[t]
\small
\vspace{-1.5em}
\begin{center}
\begin{tabular}{lc}
\toprule
Method & Accuracy \\
\midrule
Stack-NMN \cite{hu2018explainable} & 56.5\% \\
SLR \cite{yu2017joint} & 57.7\% \\
MAttNet \cite{yu2018mattnet} & 60.9\% \\
\midrule
GroundeR \cite{rohrbach2016grounding} & 61.7\% \\
GroundeR + LCGN w/o txt & 65.0\% \\
GroundeR + LCGN w/ static $w_{j,i}$ & 71.4\% \\
GroundeR + LCGN (ours) & \textbf{74.8\%} \\
\bottomrule
\end{tabular}
\vspace{-0.7em}
\end{center}
\caption{Performance on the CLEVR-Ref+ dataset for REF.}
\label{tab:clevr_refplus}
\vspace{-1.5em}
\end{table}

In previous works, SLR \cite{yu2017joint} and MAttNet \cite{yu2018mattnet} are specifically designed for the REF task. SLR jointly trains an expression generation model (speaker) and an expression comprehension model (listener), and MAttNet relies on modular structure for subject, location and relation comprehension. While Stack-NMN \cite{hu2018explainable} is also a generic approach that is applicable to both the VQA task and the REF task, the major contribution of Stack-NMN is to construct an explicit step-wise inference procedure with compositional modules, and it relies on hand-designed module structures and local appearance-based scene representations. On the other hand, our work augments the scene representation with rich relational context. We show that our approach outperforms Stack-NMN on both the VQA and the REF tasks.

\section{Conclusion}

In this work, we propose Language-Conditioned Graph Networks (LCGN), a generic approach to language-based reasoning tasks such VQA and REF. Instead of building task-specific inference procedures, our LCGN model constructs rich context-aware \textit{representations} of the scene through iterative message passing. Experimentally, we show that the context-aware representations from our LCGN model can improve over the local appearance-based representations across various types of local features and multiple datasets, and it is crucial for the message passing procedure to depend on the language inputs.

\noindent\textbf{Acknowledgements.} This work was partially supported by the Berkeley AI Research, the NSF and DARPA XAI.

\clearpage

{\small
\bibliographystyle{ieee_fullname}
\bibliography{lcgn_bib}
}

\clearpage

\appendix

\setcounter{table}{0}
\renewcommand{\thetable}{\Alph{section}.\arabic{table}}
\setcounter{figure}{0}
\renewcommand{\thefigure}{\Alph{section}.\arabic{figure}}
\setcounter{equation}{0}
\renewcommand{\theequation}{\Alph{section}.\arabic{equation}}

\section{Implementation details}

In our implementation, we use $d_{txt} = 512$ as the dimensionality of the textual vectors (such as $h_s$, $q$, and $c_t$), and $d_{ctx} = 512$ as the dimensionality of the context features $x^{ctx}_i$ of each entity $i$.

On the GQA dataset, we first reduce the dimensionality of the input local features $x^{loc}_i$ (convolutional grid features, object detection features or GT objects and attributes in Table~2 of the main paper) to the same dimensionality $d_{loc}=512$ with a single fully-connected layer (without non-linearity). During training, we train with a sigmoid cross entropy loss and use the Adam optimizer \cite{kingma2014adam} with a batch size of $128$ and a learning rate of $3 \times 10^{-4}$.

\begin{table}[b]
\vspace{-1em}
\begin{center}
\begin{tabular}{ccc}
\toprule
Parameter & Shape & Shared across $t$ \\
\midrule
\multicolumn{3}{c}{(textual command extraction)} \\
$W_1$ & $1 \times d_{txt}$ & yes \\
$W_2^{(t)}$ & $d_{txt} \times d_{txt}$ & no \\
$W_3^{(t)}$ & $d_{txt} \times d_{txt}$ & yes \\
\midrule
\multicolumn{3}{c}{(language-conditioned message passing)} \\
$W_4$ & $d_{ctx} \times d_{loc}$ & yes \\
$W_5$ & $d_{ctx} \times d_{ctx}$ & yes \\
$W_6$ & $d_{ctx} \times (d_{loc}+2d_{ctx})$ & yes \\
$W_7$ & $d_{ctx} \times (d_{loc}+2d_{ctx})$ & yes \\
$W_8$ & $d_{ctx} \times d_{txt}$ & yes \\
$W_9$ & $d_{ctx} \times (d_{loc}+2d_{ctx})$ & yes \\
$W_{10}$ & $d_{ctx} \times d_{txt}$ & yes \\
$W_{11}$ & $d_{ctx} \times 2d_{ctx}$ & yes \\
$W_{12}$ & $d_{loc} \times (d_{loc}+d_{ctx})$ & yes \\
\midrule
\multicolumn{3}{c}{(the single-hop answer classifier for VQA)} \\
$W_{13}$ & $1 \times d_{loc}$ & n/a \\
$W_{14}$ & $d_{loc} \times d_{txt}$ & n/a \\
$W_{15}$ & $d_{ans} \times 512$ & n/a \\
$W_{16}$ & $512 \times (d_{loc}+d_{txt})$ & n/a \\
\midrule
\multicolumn{3}{c}{(GroundeR for REF)} \\
$W_{17}$ & $1 \times d_{loc}$ & n/a \\
$W_{18}$ & $d_{loc} \times d_{txt}$ & n/a \\
$W_{19}$ & $4 \times d_{loc}$ & n/a \\
\bottomrule
\end{tabular}
\end{center}
\vspace{-0.5em}
\caption{The parameter shapes in our LCGN model. All parameters are shared across different time steps $t$, except for $W_2^{(t)}$.}
\label{tab:parameter_shape}
\end{table}

On the CLEVR dataset and the CLEVR-Ref+ dataset, we first apply a small two-layer convolutional network on the ResNet-101-C4 features to output a $14 \times 14 \times 512$ feature map, so that the feature dimensionality at each location on the feature map is also reduced to $d_{loc}=512$. We use the Adam optimizer \cite{kingma2014adam} with a batch size of $64$ and a learning rate of $10^{-4}$. On the CLEVR dataset, we train with a softmax loss for answer classification. On the CLEVR-Ref+ dataset, we train with a softmax loss to select the target location $p$ and an L2 loss (\ie mean squared error loss) for the bounding box offset $u$.

To facilitate reasoning about spatial relations such as ``left'' and ``right'', we also add spatial information to the local features. On the GQA dataset, when using object detection features or GT objects and attributes, we concatenate the local features with the bounding box coordinates of the corresponding objects. When using convolutional grid features (on GQA, CLEVR and CLEVR-Ref+), for each convolutional grid location $(h, w)$, we concatenate the sinusoidal positional encoding \cite{vaswani2017attention} of $h$ and $w$ to the convolutional channel output at $(h, w)$.

The shapes of the parameters in our LCGN model are shown in Table~\ref{tab:parameter_shape}. All our models are trained using a single Titan Xp GPU.

\section{Quantitative analysis on edge weights}

We perform quantitative analysis on the learned edge weights $\{w_{j,i}\}$, and measure how much they vary across different questions on the same image using the CLEVR dataset (where all images have exactly 10 associated questions). For each receiver node $i$, we associate it with a max-connected sender node $j^*=\arg\max_j \{w_{j,i}\}$. Then, we count for each receiver node $i$ in the image how many unique $j^*$ there are (across the 10 questions) -- this number would be between 1 and 10 for each image; the higher number, the more variance in $w_{j,i}$ across questions. On average, each source node $i$ is connected to 6.396 unique $j^*$ across the 10 questions on the same image, showing that the learned edge weights $\{w_{j,i}\}$ are largely dependent on the input questions.

\section{Additional visualization examples}

\begin{figure*}[t]
\centering
\small
\begin{tabularx}{\linewidth}{*{6}{c}}
~~~~~~~~input image & ~~~~~~~~~~~~~~~~~$t = 1$ & ~~~~~~~~~~~~~~~~~~~~~~~$t = 2$ & ~~~~~~~~~~~~~~~~~~~~~~$t = 3$ & ~~~~~~~~~~~~~~~~~~~~~~~$t = 4$ & ~~~~~~~~~single-hop attention $\beta_i$ \\
\hline
\end{tabularx}

\vspace{1mm}

question: \textit{are there carts near the pond?} ~ prediction: \textit{yes} ~ ground-truth: \textit{yes} \\
\includegraphics[width=\linewidth,trim={0 8mm 0 7mm},clip]{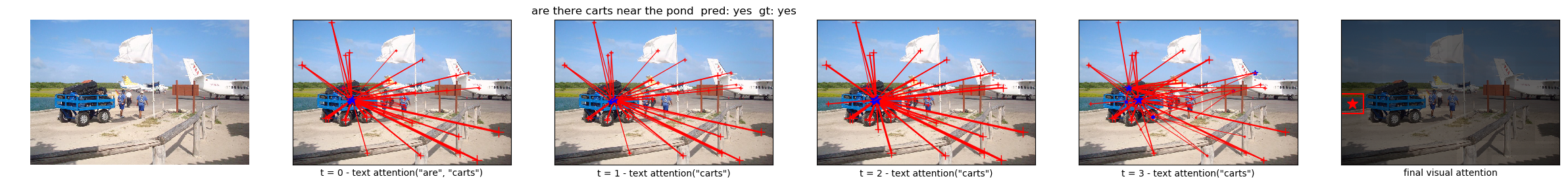} \\

question: \textit{what color is the flag?} ~ prediction: \textit{white} ~ ground-truth: \textit{white} \\
\includegraphics[width=\linewidth,trim={0 8mm 0 7mm},clip]{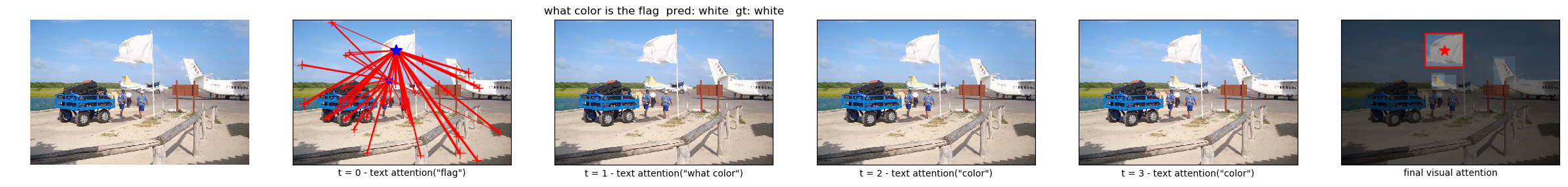} \\

question: \textit{what type of vehicle is in front of the hanging wires?} ~ prediction: \textit{train} ~ ground-truth: \textit{train} \\
\includegraphics[width=\linewidth,trim={0 8mm 0 7mm},clip]{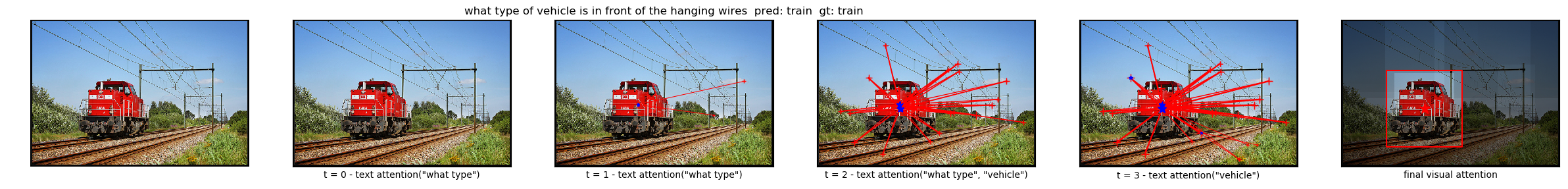} \\

question: \textit{on what does the man sit?} ~ prediction: \textit{bench} ~ ground-truth: \textit{bench} \\
\includegraphics[width=\linewidth,trim={0 8mm 0 7mm},clip]{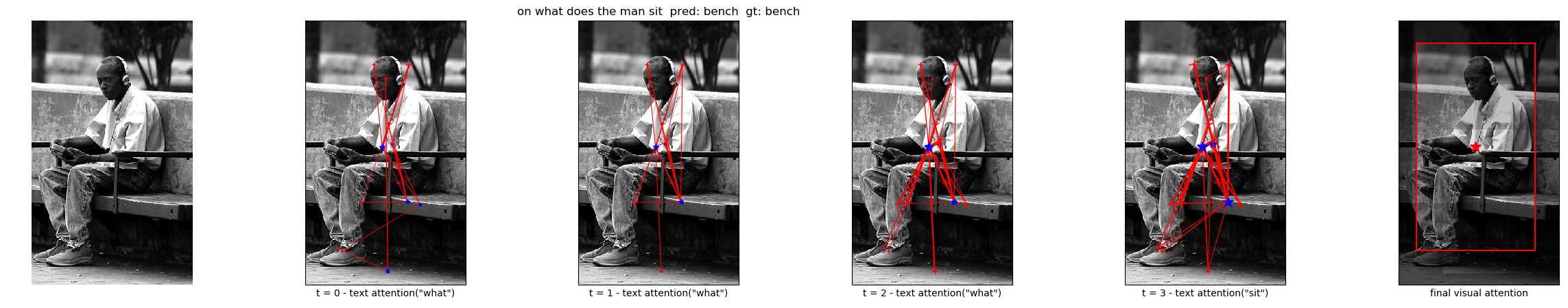} \\

question: \textit{are there both a tennis ball and a racket in the image?} ~ prediction: \textit{yes} ~ ground-truth: \textit{yes} \\
\includegraphics[width=\linewidth,trim={0 8mm 0 7mm},clip]{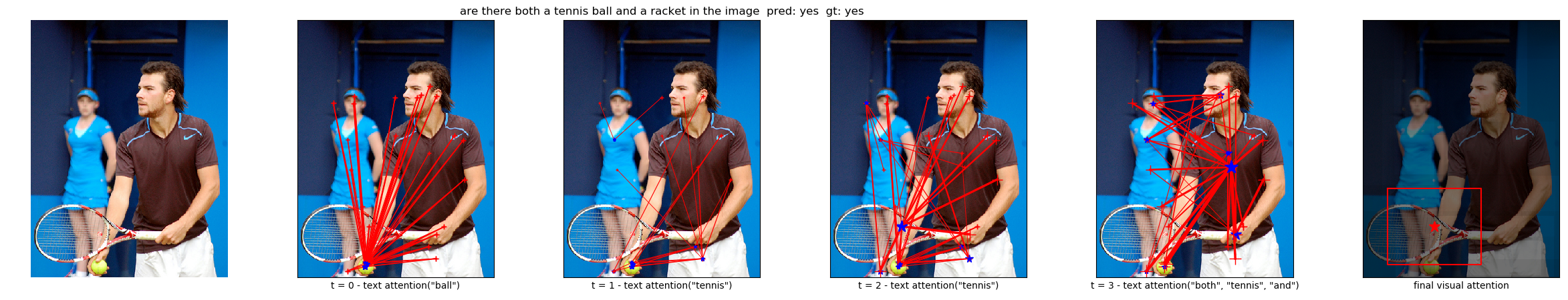} \\

question: \textit{what vehicle is on the highway?} ~ prediction: \textit{truck} ~ ground-truth: \textit{ambulance} \\
\includegraphics[width=\linewidth,trim={0 8mm 0 7mm},clip]{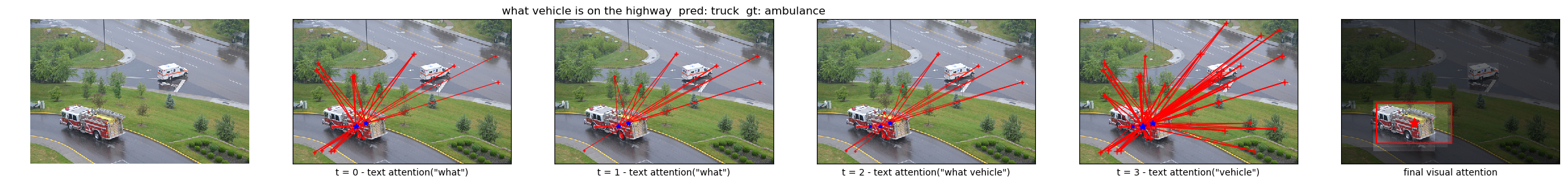} \\

question: \textit{who is holding the umbrella?} ~ prediction: \textit{woman} ~ ground-truth: \textit{lady} \\
\includegraphics[width=\linewidth,trim={0 8mm 0 7mm},clip]{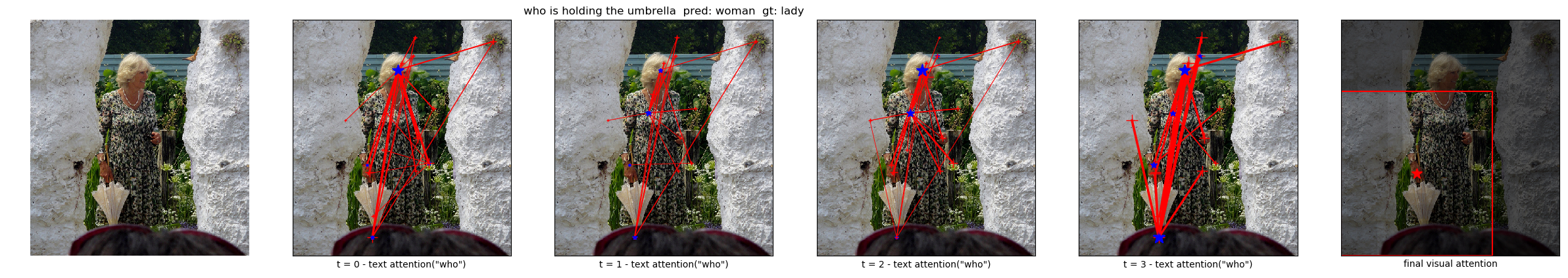} \\

\caption{Additional examples from our LCGN model on the validation split of the GQA dataset for VQA. In the middle 4 columns, each red line shows an edge $j \rightarrow i$ along the message passing paths (among the $N$ detected objects) where the connection edge weight $w^{(t)}_{j,i}$ exceeds a threshold. The \textcolor{blue}{blue} star on each line is the sender node $j$. In these example, the objects of interest receive messages from other objects through those connections with high weights (the red lines). The \textcolor{red}{red} star (along with the box) in the last column shows the object with the highest attention $\beta_i$ in the single-hop VQA classifier in Sec.~3.2 of the main paper. The last two rows show two failure examples on the GQA dataset. Some failure cases are due to ambiguity in the answers in the GQA dataset (\eg ``woman'' vs. ``lady'' in the last example).}
\label{fig:gqa_vis_supp}
\end{figure*}

\begin{figure*}[t]
\centering
\small
\begin{tabularx}{\linewidth}{*{6}{c}}
~~~~~~~~input image & ~~~~~~~~~~~~~~~~~$t = 1$ & ~~~~~~~~~~~~~~~~~~~~~~~$t = 2$ & ~~~~~~~~~~~~~~~~~~~~~~$t = 3$ & ~~~~~~~~~~~~~~~~~~~~~~~$t = 4$ & ~~~~~~~~~single-hop attention $\beta_i$ \\
\hline
\end{tabularx}

\vspace{1mm}

question: \textit{there is a small gray block ; are there any spheres to the left of it?} ~ prediction: \textit{yes} ~ ground-truth: \textit{yes} \\
\includegraphics[width=\linewidth,trim={0 8mm 0 7mm},clip]{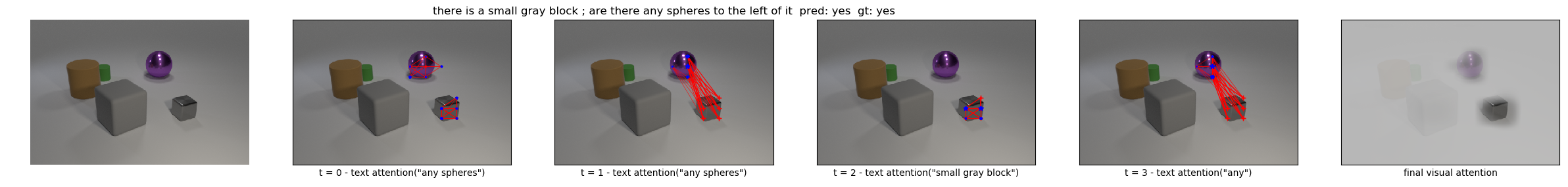} \\

question: \textit{is the purple thing the same shape as the large gray rubber thing?} ~ prediction: \textit{no} ~ ground-truth: \textit{no} \\
\includegraphics[width=\linewidth,trim={0 8mm 0 7mm},clip]{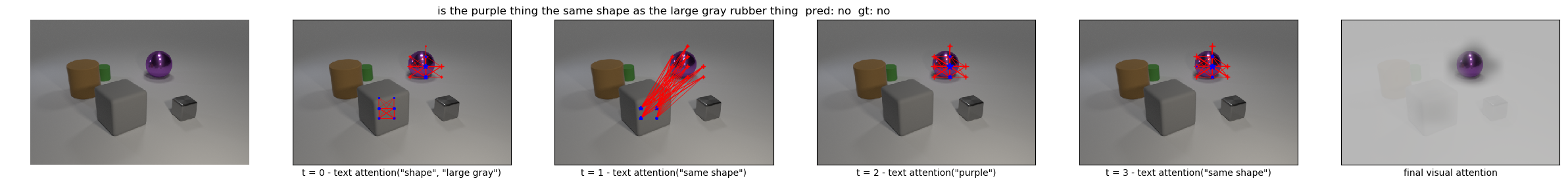} \\

question: \textit{do the large metal sphere and the matte block have the same color?} ~ prediction: \textit{yes} ~ ground-truth: \textit{yes} \\
\includegraphics[width=\linewidth,trim={0 8mm 0 7mm},clip]{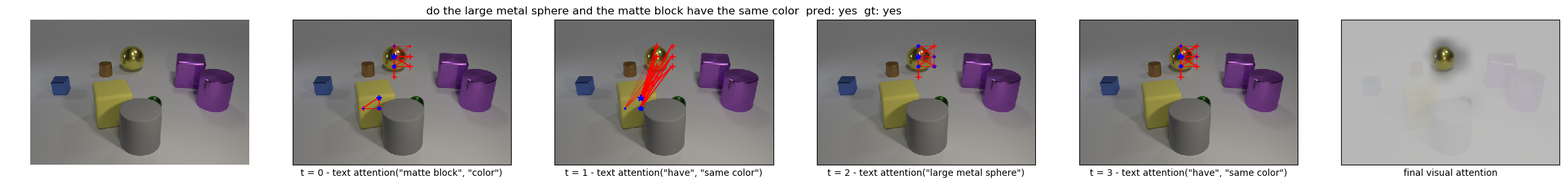} \\

question: \textit{is there anything else that has the same material as the red thing?} ~ prediction: \textit{yes} ~ ground-truth: \textit{yes} \\
\includegraphics[width=\linewidth,trim={0 8mm 0 7mm},clip]{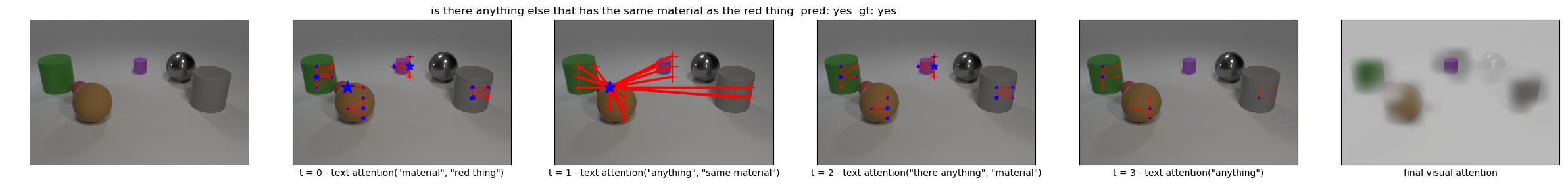} \\

question: \textit{is there any other thing that is the same color as the cylinder?} ~ prediction: \textit{no} ~ ground-truth: \textit{no} \\
\includegraphics[width=\linewidth,trim={0 8mm 0 7mm},clip]{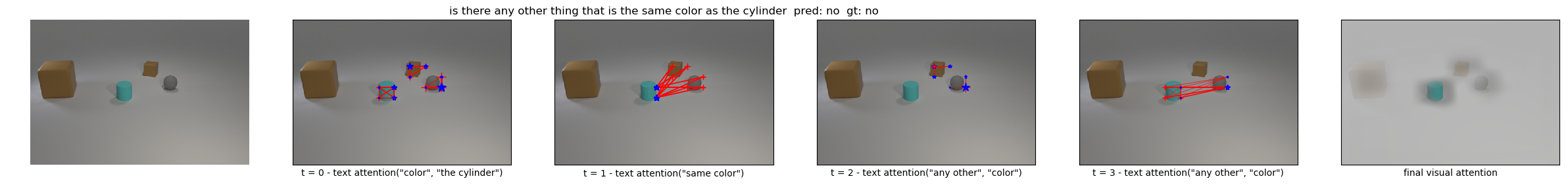} \\

question: \textit{what number of other objects are there of the same size as the gray sphere?} ~ prediction: \textit{5} ~ ground-truth: \textit{5} \\
\includegraphics[width=\linewidth,trim={0 8mm 0 7mm},clip]{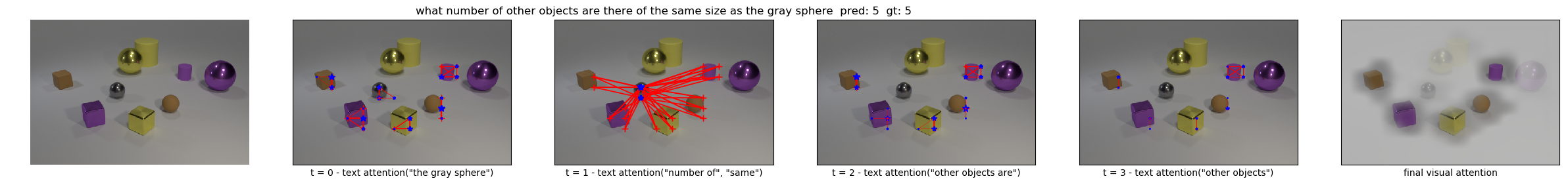} \\

question: \textit{is the number of small cylinders behind the cyan thing greater than the number of cubes that are behind the green block?} ~ prediction: \textit{yes} ~ ground-truth: \textit{no} \\
\includegraphics[width=\linewidth,trim={0 8mm 0 7mm},clip]{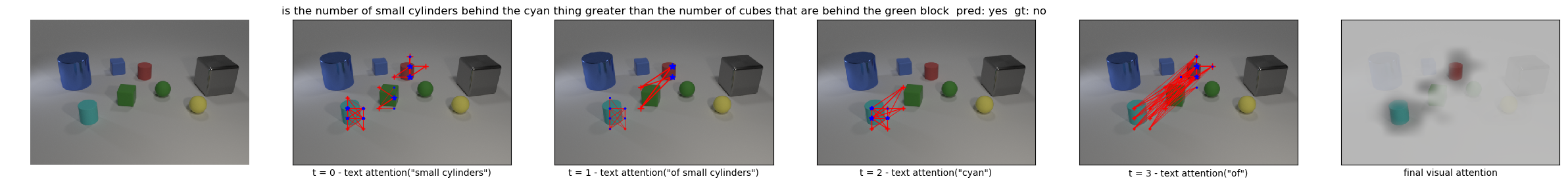} \\

question: \textit{how many other objects are the same shape as the purple metallic thing?} ~ prediction: \textit{6} ~ ground-truth: \textit{7} \\
\includegraphics[width=\linewidth,trim={0 8mm 0 7mm},clip]{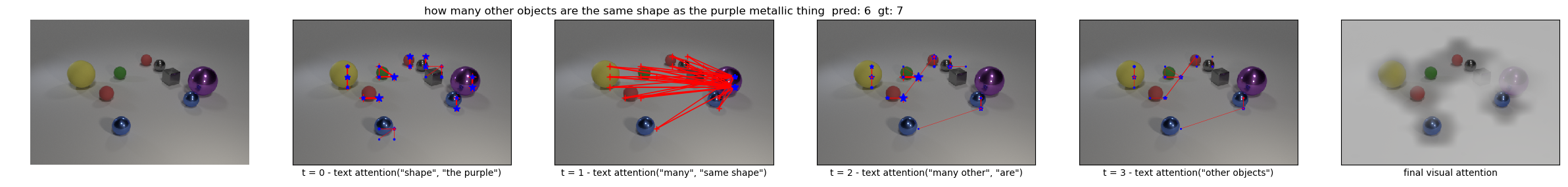} \\

\caption{Additional examples from our LCGN model on the validation split of the CLEVR dataset for VQA. The middle 4 columns show the connection edge weights $w^{(t)}_{j,i}$ similar to Figure~\ref{fig:gqa_vis_supp}, where the \textcolor{blue}{blue} stars are the sender nodes. The last column shows the attention $\beta_i$ in the single-hop VQA classifier in Sec.~3.2 of the main paper over the $N = 14 \times 14$ feature grid. In these examples, the relevant objects in the question usually first propagate messages within the convolutional grids of the same object (possibly to form an object representation from the CNN features), and then the object of interest tends to collect messages from other context objects. The last two rows show two failure examples on the CLEVR dataset.}
\label{fig:clevr_vis_supp}
\end{figure*}

\begin{figure*}[t]
\centering
\small
\begin{tabularx}{\linewidth}{*{6}{c}}
~~~~~~~~input image & ~~~~~~~~~~~~~~~~~$t = 1$ & ~~~~~~~~~~~~~~~~~~~~~~~$t = 2$ & ~~~~~~~~~~~~~~~~~~~~~~$t = 3$ & ~~~~~~~~~~~~~~~~~~~~~~~$t = 4$ & ~~~~~~~~~~bounding box output \\
\hline
\end{tabularx}

\vspace{1mm}

referring expression: \textit{any other yellow shiny objects that have the same size as the first one of the objects from front} \\
\includegraphics[width=\linewidth,trim={0 8mm 0 7mm},clip]{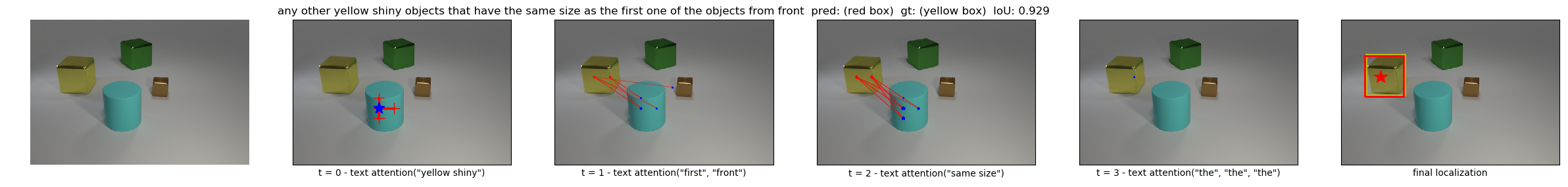} \\

referring expression: \textit{any other tiny objects that have the same material as the third one of the objects from left} \\
\includegraphics[width=\linewidth,trim={0 8mm 0 7mm},clip]{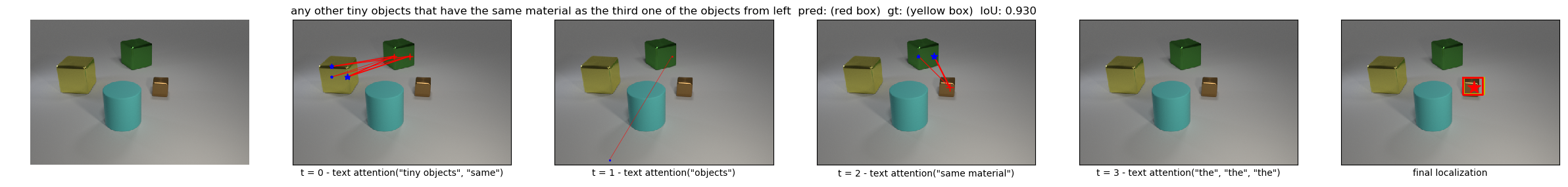} \\

referring expression: \textit{the second one of the things from left} \\
\includegraphics[width=\linewidth,trim={0 8mm 0 7mm},clip]{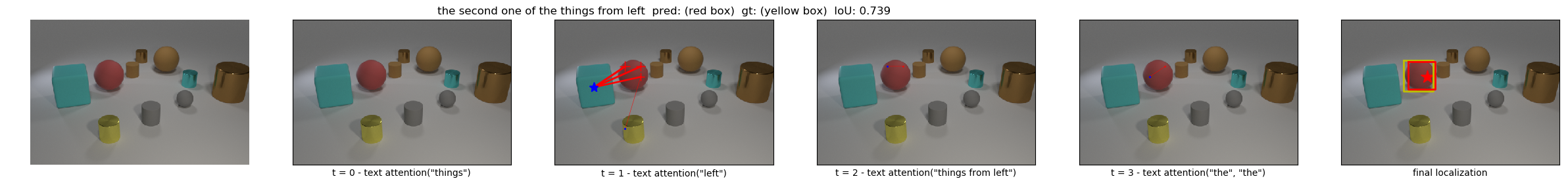} \\

referring expression: \textit{any other matte things that have the same shape as the first one of the red metal things from right} \\
\includegraphics[width=\linewidth,trim={0 8mm 0 7mm},clip]{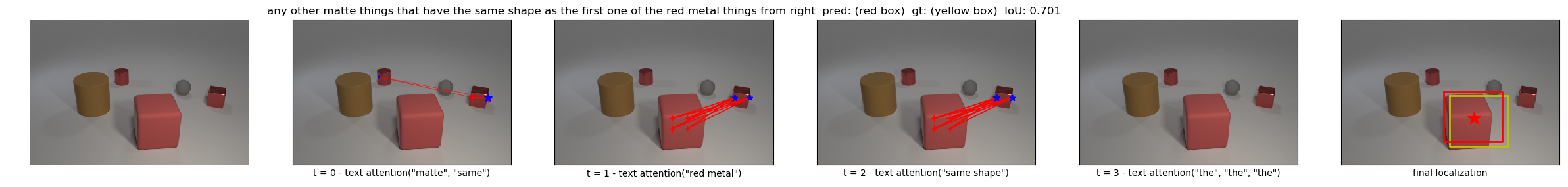} \\

referring expression: \textit{the first one of the things from front that are on the right side of the first one of the purple spheres from front} \\
\includegraphics[width=\linewidth,trim={0 8mm 0 7mm},clip]{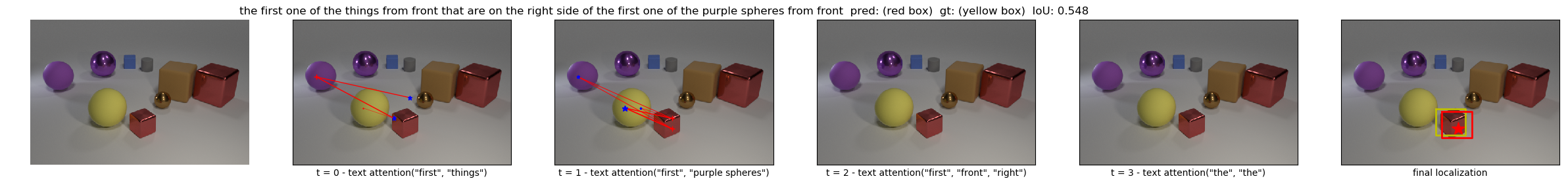} \\

referring expression: \textit{the second one of the shiny objects from front} \\
\includegraphics[width=\linewidth,trim={0 8mm 0 7mm},clip]{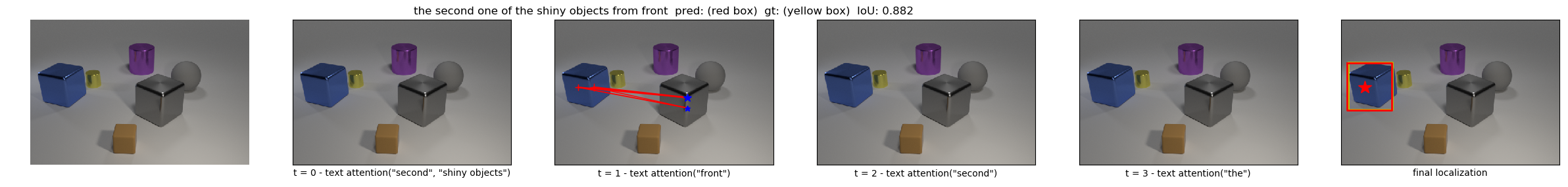} \\

referring expression: \textit{any other matte things of the same shape as the fifth one of the rubber things from right} \\
\includegraphics[width=\linewidth,trim={0 8mm 0 7mm},clip]{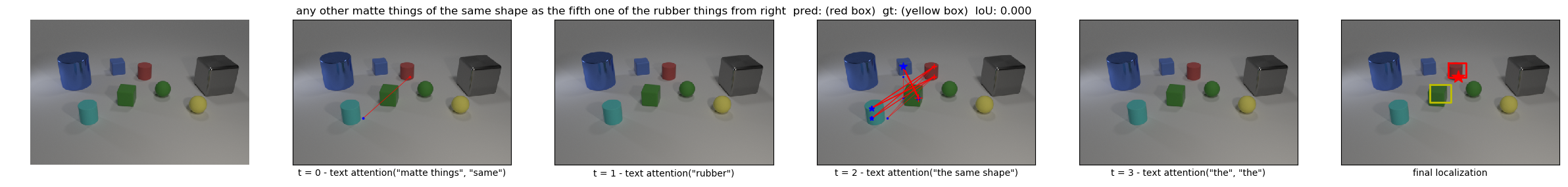} \\

referring expression: \textit{look at sphere that is right of the first one of the things from front; the second one of the objects from right that are in front of it} \\
\includegraphics[width=\linewidth,trim={0 8mm 0 7mm},clip]{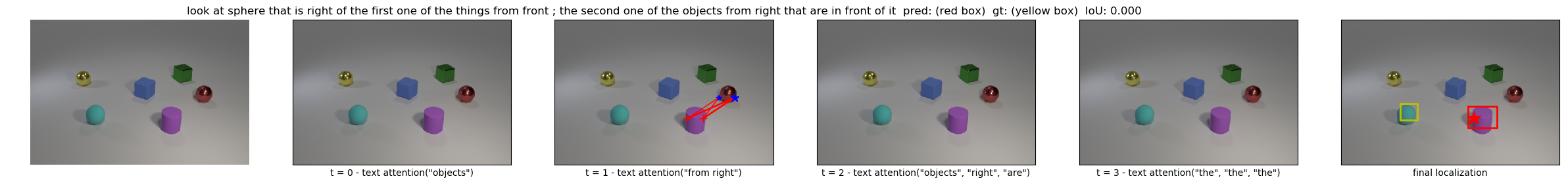} \\

\caption{Additional examples from our LCGN model on the validation split of the CLEVR-Ref+ dataset for REF. The middle 4 columns show the connection edge weights $w^{(t)}_{j,i}$ similar to Figure~\ref{fig:gqa_vis_supp}, where the \textcolor{blue}{blue} stars are the sender nodes. The last column shows the selected target grid location $p$ on the $N = 14 \times 14$ spatial grid (the \textcolor{red}{red} star) in the GroundeR model in Sec.~3.2 of the main paper, along with the ground-truth (\textcolor[rgb]{.7,.7,0}{yellow}) box and the predicted box (\textcolor{red}{red} box from bounding box regression $u$ in GroundeR). In these examples, the objects of interest tend to collect messages from other context objects. The last two rows show two failure examples on the CLEVR-Ref+ dataset.}
\label{fig:clevr_refplus_vis_supp}
\end{figure*}

Figures~\ref{fig:gqa_vis_supp} and \ref{fig:clevr_vis_supp} show additional visualization examples for the VQA task on the GQA dataset and the CLEVR dataset, respectively. Figure~\ref{fig:clevr_refplus_vis_supp} shows additional examples for the REF task on the CLEVR-Ref+ dataset.

\end{document}